\documentclass[10pt,twocolumn,letterpaper]{article}

\usepackage{iccv}              

\definecolor{iccvblue}{rgb}{0.21,0.49,0.74}
\usepackage[pagebackref,breaklinks,colorlinks,allcolors=iccvblue]{hyperref}
\PassOptionsToPackage{table}{xcolor}
\usepackage{xcolor}
\usepackage{colortbl}
\usepackage{multirow}
\usepackage{graphicx}
\usepackage{amssymb}
\usepackage{pifont}

\def\paperID{***} 
\def\confName{ICCV}
\def\confYear{2025}

\title{LUT-Fuse: Towards Extremely Fast Infrared and Visible Image Fusion via Distillation to Learnable Look-Up Tables}

\author{Xunpeng Yi$^{1,*}$, Yibing Zhang$^{1,*}$, Xinyu Xiang$^{1}$, Qinglong Yan$^{1}$, Han Xu$^{2}$, Jiayi Ma$^{1,\dagger}$\\
$^1$Electronic Information School, Wuhan University, Wuhan 430072, China\\
$^2$School of Automation, Southeast University, Nanjing 210096, China\\
\tt\small 
\{yixunpeng, zhangyibing, xiangxinyu, qinglong\_yan\}@whu.edu.cn, \\
\tt\small xu\_han@seu.edu.cn, jyma2010@gmail.com
}

\begin{document}
\maketitle

\begingroup
\renewcommand\thefootnote{}
\footnotetext{
\makebox[0pt][l]{
  \begin{minipage}[t]{\dimexpr\linewidth}
  \raggedright
  $^\dagger$ Corresponding author. \\
  $^*$ These authors contributed equally to this work.
  \end{minipage}
}
}
\endgroup

\begin{abstract}
Current advanced research on infrared and visible image fusion primarily focuses on improving fusion performance, often neglecting the applicability on real-time fusion devices. In this paper, we propose a novel approach that towards extremely fast fusion via distillation to learnable lookup tables specifically designed for image fusion, termed as LUT-Fuse. Firstly, we develop a look-up table structure that utilizing low-order approximation encoding and high-level joint contextual scene encoding, which is well-suited for multi-modal fusion. Moreover, given the lack of ground truth in multi-modal image fusion, we naturally proposed the efficient LUT distillation strategy instead of traditional quantization LUT methods. By integrating the performance of the multi-modal fusion network (MM-Net) into the MM-LUT model, our method achieves significant breakthroughs in efficiency and performance. It typically requires less than one-tenth of the time compared to the current lightweight SOTA fusion algorithms, ensuring high operational speed across various scenarios, even in low-power mobile devices. Extensive experiments validate the superiority, reliability, and stability of our fusion approach. The code is available at https://github.com/zyb5/LUT-Fuse.
\end{abstract}

\section{Introduction}
Image fusion represents a critical research area within the domain of digital image processing~\cite{liu2023multi, xu2022rfnet, zhu2022clf, he2023degradation, wang2024terf, qi2024dmfuse}. Single-modal imaging systems are inherently limited in their ability to capture complete scene information, resulting in constrained information representation. In contrast, multi-modal imaging systems, by integrating complementary data sources, achieve more comprehensive scene characterization~\cite{yi2025artificial, xu2021drf, liu2024promptfusion}. A prominent example of multi-modal image fusion is infrared and visible image fusion (IVIF). Visible images offer detailed reflection-based visual information, while infrared images provide thermal radiation information reflecting temperature variations in the scene~\cite{ma2019infrared, ma2020ddcgan, huang2022reconet}. This synergistic utilization of multiple modalities has significant applications across various fields, including industrial inspection systems, autonomous vehicle navigation, and military night vision technologies.

\begin{figure}[!t]
\centering
\includegraphics[width=0.98\linewidth]{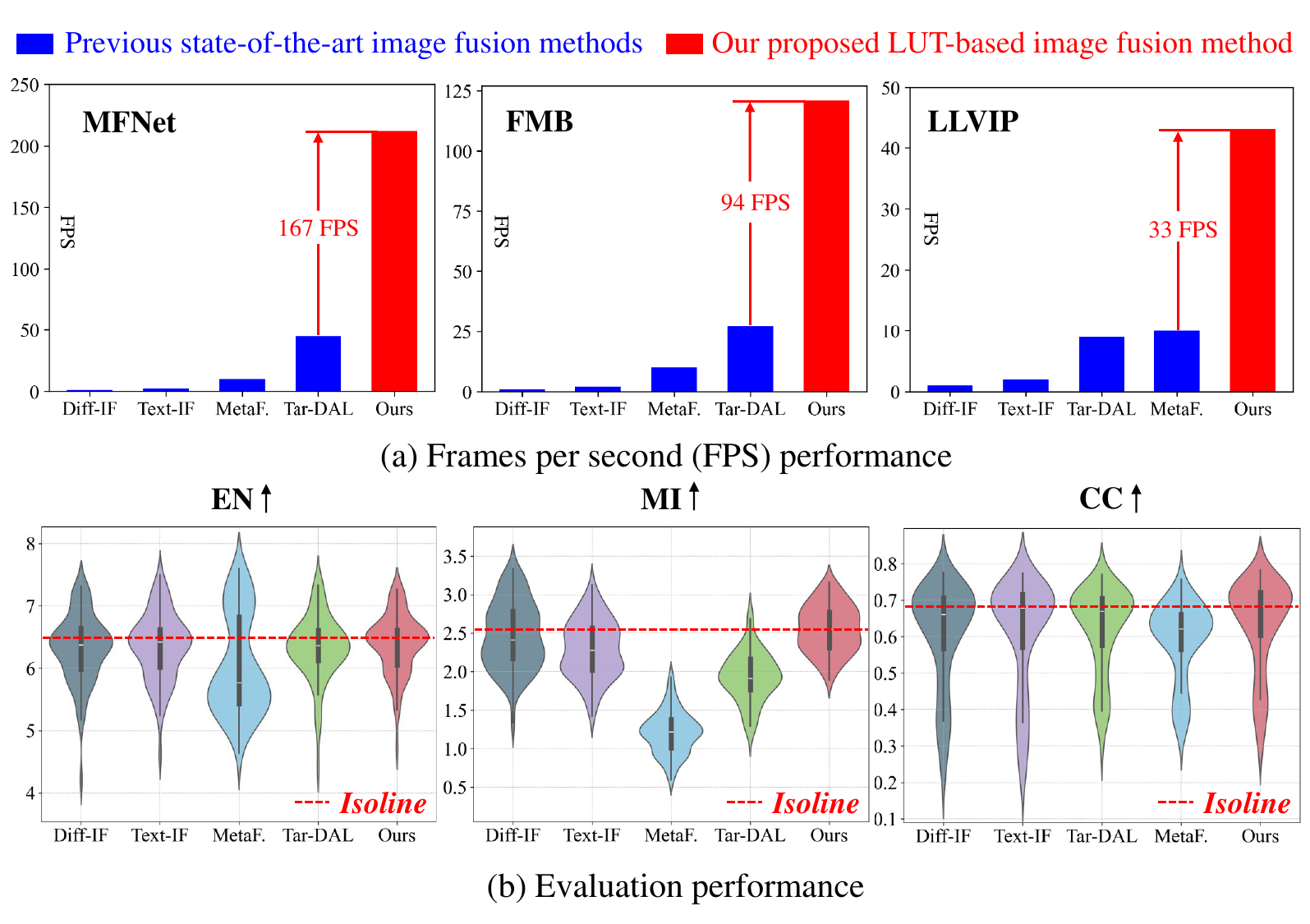}
\vspace{-0.1in}
\caption{(a) Our proposed LUT-Fuse achieves real-time FPS performance compared to state-of-the-art methods, demonstrating superior efficiency, the improvements are over \textbf{100 FPS}. (b) LUT-Fuse delivers leading or comparable image fusion performance. }
\label{fig:1-1}
\end{figure}

The application-specific requirements of multi-modal image fusion tasks necessitate the simultaneous achievement of high computational efficiency and superior fusion performance across diverse operational scenarios~\cite{zhang2021sdnet}. This dual requirement stems from two fundamental constraints: the computational limitations inherent in deployed devices and the downstream task performance demands of practical fusion applications. In recent years, there has been a surge in the development of advanced architectural frameworks, particularly Transformer-based~\cite{zhao2023cddfuse, yi2024text} and diffusion-based models~\cite{yi2024diff, zhao2023ddfm}, specifically designed to improve fusion performance. These innovative approaches have achieved remarkable breakthroughs, consistently demonstrating superior performance across diverse application scenarios. However, a critical limitation in current research is that these studies predominantly focus on fusion performance metrics while significantly overlooking real-time processing considerations. Notably, the majority of these proposed methods fail to achieve real-time operational efficiency, even when implemented on state-of-the-art GPU hardware platforms. Even though some methods claim to achieve real-time operation, they primarily rely on the design of lightweight structures~\cite{zhang2021sdnet, liu2022target, zhang2023real}. Moreover, their real-time operation scenarios are strictly limited, achieving only quasi-real-time performance in partial scenarios. Therefore, addressing real-time challenges in IVIF tasks has become an imperative research priority.

Look-up tables (LUTs) represent a widely adopted technology in data storage systems, enabling rapid retrieval of corresponding outputs through high-speed query mechanisms~\cite{meher2010lut, zeng2020learning}. This characteristic offers a promising solution for addressing computational efficiency challenges in image fusion tasks. Nevertheless, the direct transformation of fusion tasks into lookup-based operations faces two fundamental limitations: (1) \textit{The inherent absence of ground truth in fusion tasks prevents explicit deployment of LUT-based solutions.} (2) \textit{Conventional quantization non-learnable approaches yield LUTs~\cite{jiang2023meflut} with limited generalization capability and suboptimal fusion performance.}

To address these issues, we propose a novel approach focusing on extremely fast fusion via distillation to learnable look-up tables designed for image fusion, termed as LUT-Fuse. Firstly, leveraging the low-order approximation and learnable scene relationships, we develop a comprehensive framework of MM-LUT, comprising zeroth-order and first-order components and the learnable scene context component, specifically designed to generate representative fusion look-up elements. Secondly, given the absence of ground truth in MMIF tasks, the proposed approach naturally incorporates fusion performance through an efficient MM-LUT distillation paradigm, effectively transferring the multi-modal fusion network (MM-Net) prior capabilities to MM-LUT. Also, the proposed LUT-Fuse employs the learnable MM-LUT designed for image fusion as the student model, enabling ultra-efficient fusion during inference. Consequently, LUT-Fuse successfully achieves a dual optimization, simultaneously delivering both real-time processing capability and superior fusion performance, as in Fig.~\ref{fig:1-1}.

Overall, our contributions can be summarized as follows:
\begin{itemize}
    \item Towards extremely fast infrared and visible image fusion, we propose the learnable multi-modal fusion look-up tables, which contains low-order approximation encoding, and high-level joint contextual scene encoding as the look-up elements. These properties closely related to fusion can ensure the effectiveness of look-up tables.
    \item Given the inherent absence of ground truth in multi-modal image fusion tasks, we adopt the efficient MM-LUT distillation paradigm as a natural solution. This effectively transfers the superior fusion capabilities from the multi-modal fusion network to MM-LUT model. Through iterative optimization via gradient descent, MM-LUT is refined to achieve optimal fusion performance while maintaining extremely high efficiency.
    \item To the best of our knowledge, this is the first time that efficient distillation and learnable look-up tables have been used in multi-modal image fusion, achieving real-time performance even on low-power mobile devices. It requires only about one-tenth of the computational time of most lightweight fusion algorithms while maintaining competitive performance.
\end{itemize}

\section{Related Work}

\subsection{Advanced Deep Learning-based Methods}
Image fusion has made significant progress since the development of deep learning. In the initial developmental phase, auto-encoder-based architectures~\cite{li2018densefuse, duffhauss2022fusionvae} dominated the field, typically undergoing pre-training on extensive image datasets before implementing fusion through carefully crafted, manually designed strategies. Subsequently, end-to-end trainable fusion networks utilizing Convolutional Neural Networks are proposed~\cite{li2018densefuse, xu2020fusiondn}. U2Fusion~\cite{xu2020u2fusion} implements densely connected network combined with lifelong learning mechanisms to accomplish unified image fusion across diverse scenarios. To further improve the performance, the Transformer-based and diffusion-based methods are introduced. CDDFuse~\cite{zhao2023cddfuse} employs a Transformer-based architecture integrated with invertible neural network methodologies. Moreover, Diff-IF~\cite{yi2024diff} accomplishes high-quality multi-modal image fusion by leveraging generative diffusion models. Furthermore, substantial advancements have been achieved in semantic-aware and degradation-robust image fusion methodologies, demonstrating superior fusion performance across various challenging scenarios. Text-IF~\cite{yi2024text} utilizes text-based modulation mechanisms, integrated with high-quality restored images, to accomplish degradation-aware fusion. Despite their impressive outcomes, these methods fail to satisfy the stringent real-time fusion demands required by most practical applications.

\begin{figure*}[!t]
\centering
\includegraphics[width=0.99\linewidth]{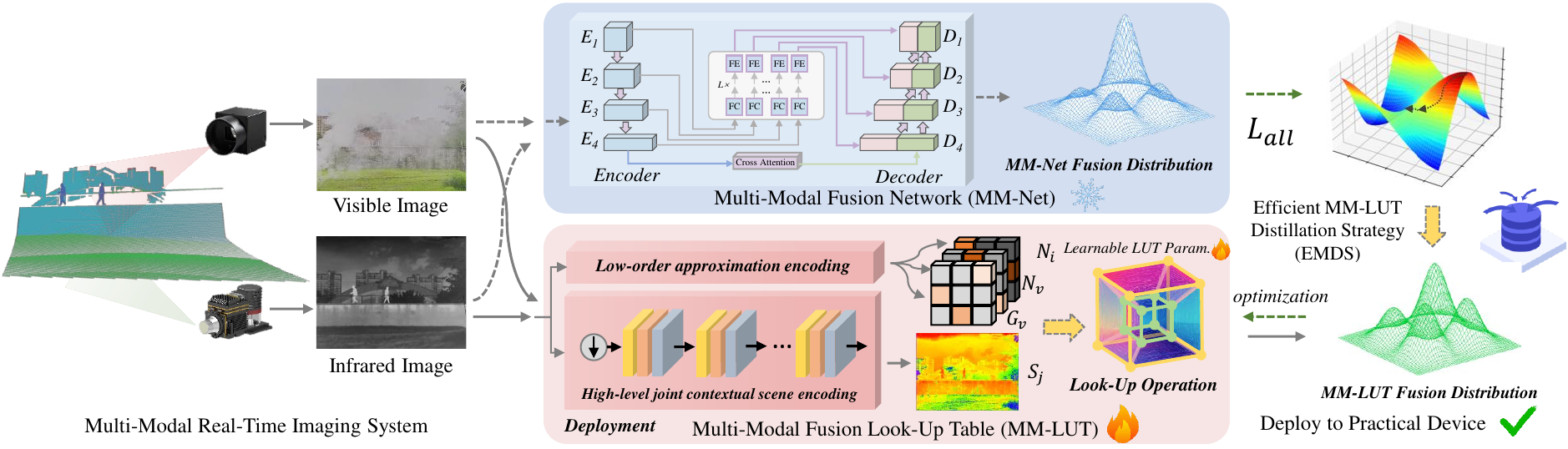}
\vspace{-0.1in}
\caption{The framework of LUT-Fuse. It consists of MM-Net and MM-LUT. MM-Net provides powerful fusion capabilities to guide the learning of MM-LUT, while MM-LUT designed for extremely fast fusion is suitable for practical deployment.}
\label{fig:3-1}
\end{figure*}

\subsection{Real-Time Deep Learning-based Methods}
Considering the efficiency limitations of computing platforms, some fusion methods for real-time operation have been proposed. However, the majority of these approaches rely primarily on the development of lightweight network to achieve their goals. In the early time, IFCNN~\cite{zhang2020ifcnn} uses several stacked convolutional layers to achieve various image fusions and is one of the representative algorithms for high-speed fusion in deep learning. Subsequently, SDNet~\cite{zhang2021sdnet} achieves fast fusion through squeeze-and-decomposition and dual-branch lightweight convolutional layers. Similarly, Tar-DAL~\cite{liu2022target} utilizes concatenation operation and compact generator architectures to achieve fusion. Recently, APWNet~\cite{zhang2023real} has employed lightweight-optimized convolutional layers as its core fusion architecture, complemented by task-specific guidance from downstream applications to enhance performance. 

These lightweight methods exhibit quasi-real-time performance in limited scenarios but struggle to sustain consistent real-time efficacy across diverse environments. The evolution of imaging technologies has intensified demands for high-resolution image fusion, driving the development of universally adaptable real-time solutions.

\section{Methodology}
In this section, we initially present the comprehensive workflow of our proposed methodology, as illustrated in Fig.~\ref{fig:3-1}. Subsequently, we provide a detailed exposition of the developed learnable multi-modal fusion look-up table and efficient MM-LUT distillation. The section concludes with a thorough specification of loss functions.

\subsection{Overall Structure} 
Leveraging the characteristics of infrared and visible image fusion tasks, we have developed an innovative multi-modal fusion look-up table (MM-LUT) architecture, which can integrate fusion capabilities from existing pre-trained multi-modal fusion networks (MM-Net). MM-LUT employs low-order approximation techniques and high-level contextual scene encoding to construct a comprehensive representation for look-up elements. Furthermore, it implements an optimization approach for MM-LUT, effectively replacing the conventional quantization approach. Given inherent absence of ground truth in MMIF, our MM-LUT distillation solution emerges as a particularly natural methodological choice, effectively addressing this fundamental limitation.

\subsection{Multi-Modal Fusion Look-Up Table}
MM-LUT architecture is systematically decomposed into two complementary components: low-order approximation and learnable higher-level joint contextual scene encoding, which collectively address diverse information aggregation requirements in multi-modal image fusion applications. 

\noindent
\textbf{Low-Order Approximation Encoding (LAE). } Inspired by the approximation principles of Taylor expansion, we decompose multi-modal image fusion into distinct hierarchical levels, each corresponding to different orders of fusion operations. Given the inherent perceptual bias towards low-order signal variations, our method emphasizes the utilization of zeroth-order and first-order components. 

\begin{figure}[t]
\centering
\includegraphics[width=0.99\linewidth]{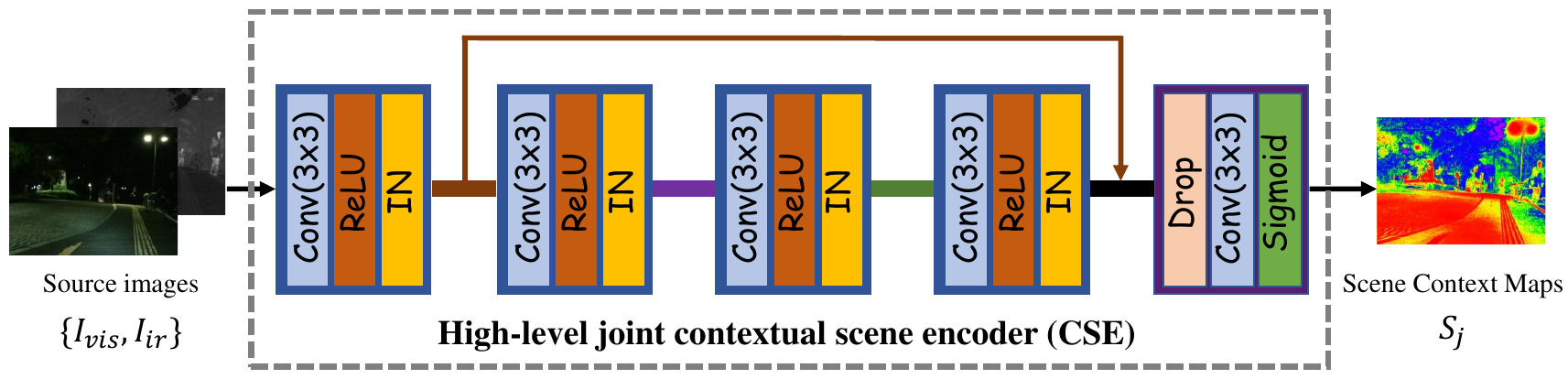}
\vspace{-0.1in}
\caption{The architectures of our high-level joint contextual scene encoder network.}
\label{fig:3-2}
\end{figure}

In the infrared and visible image fusion task, infrared images $I_{ir} \in \mathbb{R}^{H\times W}$ primarily contribute zeroth-order information, specifically intensity, denoted as $N_{i} = I_{ir}$, which effectively represents salient thermal radiation objects. Visible images $I_{vis}\in \mathbb{R}^{3 \times H\times W}$ offer both comprehensive spectral intensity characteristics and detailed texture patterns. To optimize computational efficiency, we strategically approximate these low-order features as zeroth-order (intensity) and first-order (gradient) information components, denoted as $N_{v} = I_{vis}$ and $G_{v} = \nabla_{grad} (I_{vis})$. $\nabla_{grad}$ is the first order derivative operator.

\noindent
\textbf{High-Level Joint Contextual Scene Encoding (CSE). } Although low-order information can be acquired with minimal computational overhead, it suffers from inherent representational limitations. To mitigate this constraint without compromising computing efficiency, we introduce a learnable high-level joint contextual scene encoder $\Phi_{s}$, expressed as:
\begin{equation}
\label{eq1}
S_{j} = \Phi_{s} (I_{ir}, I_{vis}).
\end{equation}

This adaptive encoding intelligently extracts look-up elements via optimized learning strategies, thereby enhancing performance. In detailed, it consists of five convolutional blocks with the kernel size of $3 \times 3$, as illustrated in Fig.~\ref{fig:3-2}.

\begin{figure}[!h]
\centering
\includegraphics[width=0.99\linewidth]{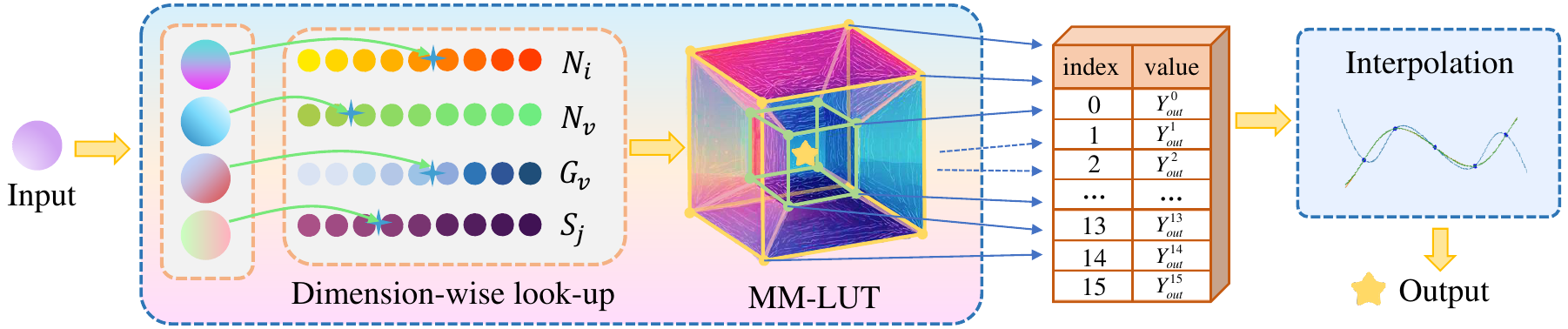}
\vspace{-0.1in}
\caption{The look-up operation and architecture of MM-LUT.}
\label{fig:LUT opera}
\end{figure}

\noindent
\textbf{Multi-Modal Look-Up Operation. } To optimize the efficiency of system integration, we have implemented a strategy that transforms large-scale computational tasks into look-up table operations, as in Fig.~\ref{fig:LUT opera}. Building upon the establishment of low-order approximation encoding and high-level joint contextual scene encoding, we further developed a LUT (towards $N_{i}(x, y)$, $N_{v}(x, y)$, $G_{v}(x, y)$ and $S_{j}(x, y)$, $x \in [1, H], y \in [1, W]$) for multi-modal image fusion:
\begin{equation}
\label{eq2}
I_{f}^{LUT}\!(x, y) \!=\! \Psi_{LUT} (\!N_{i}\!(x, y), N_{v}\!(x, y), G_{v}\!(x, y), S_{j}\!(x, y)\!),
\end{equation}
where $I_{f}^{LUT}(x, y)$ denotes output fusion image. $\Psi_{LUT}$ is the look-up operation in the MM-LUT.

They store the fusion mapping relationships in the form of a four-dimensional lattice, where each point position is determined by a quadruple $(a, b, c, d)$. To be detailed, the quadruple is computed by the following equations: 
\begin{equation}
  \begin{aligned}
    &a=\frac{N_v(x, y)}{\mathcal{T} }, \ b=\frac{N_i(x, y)}{\mathcal{T} }, \\
    &c=\frac{G_v(x, y)}{\mathcal{T} }, \ d=\frac{S_j(x, y)}{\mathcal{T} }, 
  \end{aligned}
\end{equation}
where $\mathcal{T}$ is a hyper-parameter, denoting the max value of look-up elements divided by the setting bins. Subsequently, we applied the floor function to the positional parameters of this quadruple:
\begin{equation}
  \begin{aligned}
    k=\left\lfloor a\right\rfloor, l=\left\lfloor b\right\rfloor, 
    m=\left\lfloor c\right\rfloor, n=\left\lfloor d\right\rfloor,
  \end{aligned}
\end{equation}
where $\left\lfloor \cdot \right\rfloor$ represents the floor function. Due to the discrete nature of LUT, cubic spline interpolation is required for processing. In other words, in the MM-LUT, the intermediate values between discrete points are obtained through interpolation from the surrounding points:
\begin{equation}
\label{eq3}
Y _{\text {out}}^{(a, b, c, d)} = \sum_{h\in \mathcal{D}} \sum_{p\in \mathcal{D}} \sum_{q\in \mathcal{D}} \sum_{r\in \mathcal{D}} w Y_{\text {out}}^{(k+h, l+p, m+q, n+r)},
\end{equation}
\begin{equation}
\label{eq4}
w \!= \!(1 - o_v)^{1-a} o_v^a (1 - o_g)^{1-b} o_g^b (1 - o_s)^{1-c} o_s^c (1 - o_i)^{1-d} o_i^d,
\end{equation}
where $o_v=I_v(k, l, m, n)-I_v(a, b, c, d)$, $o_g=G_v(k, l, m, n)-G_v(a, b, c, d)$, $o_s=S_j(k, l, m, n)-S_j(a, b, c, d)$, and $o_i=I_i(k, l, m, n)-I_i(a, b, c, d)$ are weighted parameters. $\mathcal{D}=\{0,1\}$. $v, g, s, i$ are four-dimensional lookup elements.

\subsection{Efficient MM-LUT Distillation Strategy}

\textbf{Efficient Distillation.} Different from the mainstream quantized LUT approaches employed for the acceleration, our MM-LUT incorporates trainable parameters to improve the performance. Given the absence of ground truth in multi-modal image fusion, we naturally propose leveraging distillation techniques to integrate fusion capabilities $I_{f}^{T} = \theta_{fuse} (I_{ir}, I_{vis})$ from the MM-Net $\theta_{fuse}$. 

Therefore, besides incorporating the trainable high-level joint contextual scene encoder to improve the performance, MM-LUT is also parameterized all internal components of the LUTs as learnable parameters, thereby circumventing the accuracy degradation and performance limitations inherent in quantization-based methods. In this way, the proposed collaborative distillation methods transform traditional quantization strategy into optimization strategy. It can be expressed as:
\begin{equation}
I_{f}^{T}=\theta_{fuse} (I_{ir}, I_{vis}) \rightarrow I_{f}^{LUT}=\theta_{MM-LUT} (I_{ir}, I_{vis}),
\end{equation}
where $\theta_{fuse}$ denotes the fusion network. $\theta_{MM-LUT}$ represents the MM-LUT. 

\noindent
\textbf{Optimizing MM-LUT Strategy.} From this perspective, MM-LUT is conceptualized as learnable parameters that undergo optimization and refinement during the distillation. We have established a distillation loss function $L_{dist}$ to regulate it, thereby enabling effective training of the MM-LUT:
\begin{equation}
\theta_{MM-LUT} \leftarrow \theta_{MM-LUT} - \eta \cdot \frac{\partial L_{dist}}{\partial \theta_{LUT}},
\end{equation}
where $\eta$ denotes the step size for each update iteration. Considering the smoothness and monotonicity that LUTs should have, it is essential to incorporate constraints as regularization terms during the optimization to ensure stability. Ultimately, we get a compact and efficient MM-LUT that is readily deployable for extremely fast multi-modal fusion applications.

\subsection{Loss Functions}
The MM-Net serves as the foundation for achieving fundamental performance. We have employed the loss functions following the methodology outlined in~\cite{yi2024text}, thereby ensuring state-of-the-art fusion outcomes.

Regarding the efficient fusion LUT distillation, we primarily employ three loss terms, including the intensity distillation loss, structural similarity distillation loss, and two LUT-specific regularization terms, namely smoothness regularization and monotonicity regularization.

\noindent
\textbf{Intensity Distillation Loss.} To ensure that fusion outcomes of LUT-Fuse exhibit intensity values comparable to those of the advanced teacher network, we employ an intensity loss function as a constraint. It is defined as:
\begin{equation}
L_{dist-int}(I_{f}^{T},I_{f}^{LUT})=||I_{f}^{T}-I_{f}^{LUT}||_{1}.
\end{equation}

\noindent
\textbf{Structural Similarity Distillation Loss.} In order to enforce structural consistency between the fusion results of LUT-Fuse and the teacher network outputs, we apply structural similarity constraints, thereby enhancing both structural coherence and scene consistency in the fused results:
\begin{equation}
  L_{dist-ssim}=1-SSIM(I_{f}^{T},I_{f}^{LUT}).
\end{equation}

\noindent
\textbf{Smooth Regularization.} Non-smooth LUTs may induce abrupt fusion output variations between adjacent look-up indices, thereby compromising the robustness of the look-up table and potentially introducing artifacts in fusion outcomes. To address this issue, smoothness regularization incorporates an L2-norm regularization term to ensure local smoothness of the LUT elements. It can be expressed as:
\begin{equation}
\begin{aligned}
  R_{TV} &\!=\!\sum_{c \in \{v,g,s,i\}}\!\sum_{k,l,m,n} \!(||c^{O}_{k+1,l,m,n}\!-\!c^{O}_{k,l,m,n}||^2\!+\\
  &\!||c^{O}_{k,l+1,m,n}-c^{O}_{k,l,m,n}||^2+||c^{O}_{k,l,m+1,n}-c^{O}_{k,l,m,n}||^2+\\
  &||c^{O}_{k,l,m,n+1}-c^{O}_{k,l,m,n}||^2).
\end{aligned}
\end{equation}

\noindent
\textbf{Monotonicity Regularization.} Monotonic transformations preserve relative intensity consistency, thereby ensuring a more natural appearance in the fusion results. Furthermore, in practical training scenarios, the available data may not adequately cover the entire look-up space. Consequently, enforcing monotonicity enhances the generalization capability of the learned LUTs:
\begin{equation}   
  \begin{aligned}
    R_m&=\sum_{c \in \{ v,g,s,i \} }\sum_{k,l,m,n} [ g(c^{O}_{(k,l,m,n)}-c^{O}_{(k+1,l,m,n)})+ \\
    &g(c^{O}_{(k,l,m,n)}-c^{O}_{(k,l+1,m,n)}) + g(c^{O}_{(k,l,m,n)}-\\
    &c^{O}_{(k,l,m+1,n)}) + g(c^{O}_{(k,l,m,n)}-c^{O}_{(k,l,m,n+1)})  ].
  \end{aligned}
\end{equation}

Therefore, the overall loss functions for LUT-Fuse can be expressed as:
\begin{equation}
  \begin{aligned}
    L_{all}\!=\!L_{dist-int}\!+\!\lambda_{ssim}L_{dist-ssim}\!+\!\lambda_{TV}R_{TV}\!+\!\lambda_{m}R_m,
  \end{aligned}
\end{equation}
where $\lambda_{ssim}$, $\lambda_{TV}$, and $\lambda_{m}$ are the hyper parameters.

\section{Experiment}

\subsection{Implementation Details and Datasets} 
\textbf{Implementation Details.} For the LUT-Fuse consists MM-Net and MM-LUT, we first train the MM-Net as the same settings of~\cite{yi2024text}. MM-Net can adopt any advanced MMIF network. We utilize a novel fusion network, with details provided in the supplementary material. For MM-LUT, the learning rate is $5e-5$ with the AdamW optimizer. And the batch size is set to 8. The source images are cropped to $96 \times 96$. The set of hyper-parameters is $\mathcal{T}=17$, $\lambda_{ssim}=0.1$, $\lambda_{TV}=1e-4$, and $\lambda_{m}=10$. The LUT-Fuse is trained for 500 epochs. Our training experiments were conducted on GeForce RTX 3090 GPU with PyTorch framework~\cite{paszke2019pytorch}. Considering practical deployment scenarios with power constraints, we evaluated performance on the GeForce RTX 4060 Ti and NVIDIA Jetson Orin NX edge platform to assess mobile deployment feasibility.

\begin{figure*}[!t]
\centering
\includegraphics[width=0.98\linewidth]{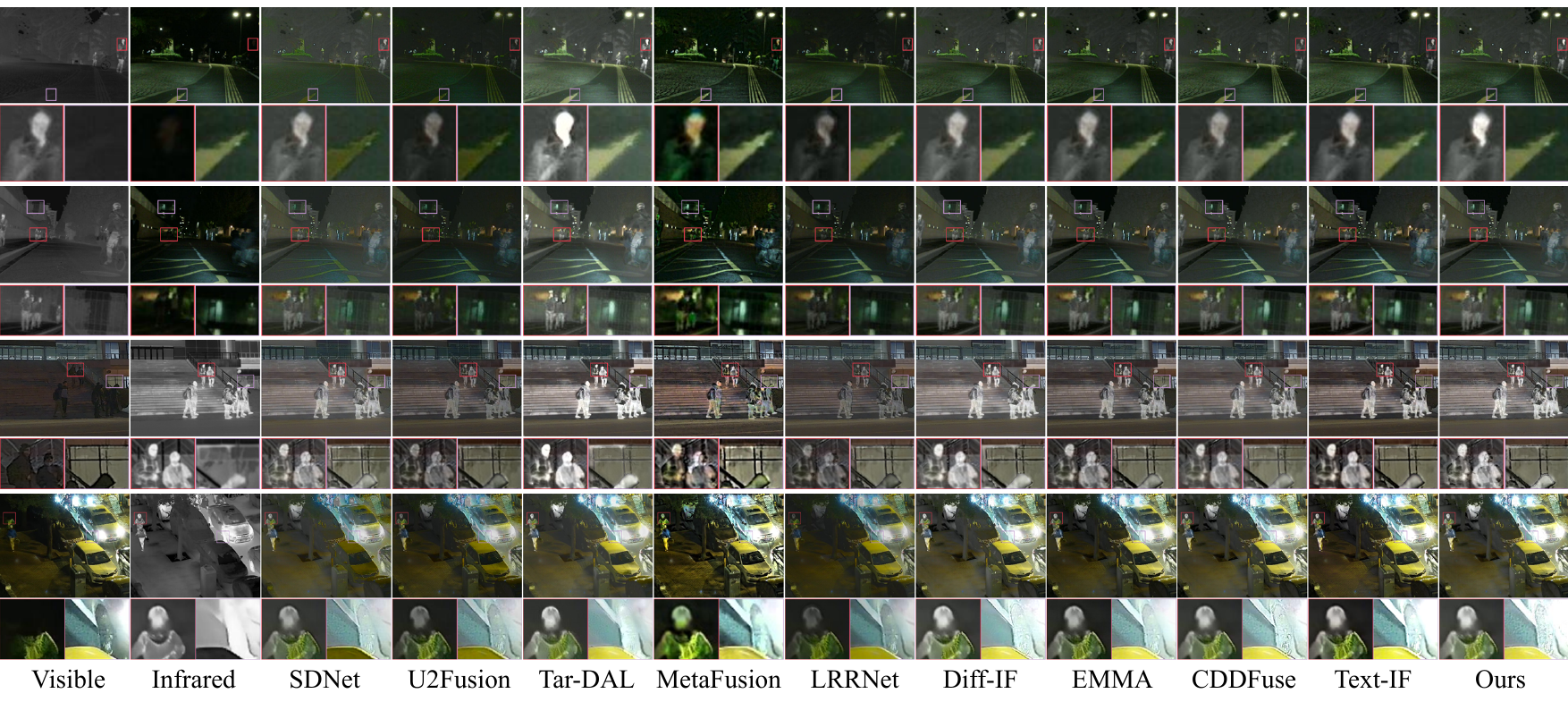}
\vspace{-0.16in}
\caption{Qualitative comparison of our proposed LUT-Fuse with the state-of-the-art multi-modal image fusion methods on MFNet, FMB, and LLVIP datasets. \textbf{Please zoom in for better viewing.}}
\label{qual_fusion}
\end{figure*}

\begin{table*}[!t]
    \renewcommand{\arraystretch}{1.15}
    \centering
    \caption{Quantitative comparison of our LUT-Fuse with existing state-of-the-art image fusion methods on the MFNet, FMB, and LLVIP datasets (\textbf{Bold}: optimal performance, \underline{underline}: second-best performance).}
    \vspace{-0.1in}
    \resizebox{0.92\textwidth}{!}{
        \begin{tabular}{c@{\,~~}|c@{\,~~~}c@{\,~~~}c@{\,~~~}c@{\,~~~}c@{\,~~~}
        |c@{\,~~~}c@{\,~~~}c@{\,~~~}c@{\,~~~}c@{\,~~~}|
        c@{\,~~~}c@{\,~~~}c@{\,~~~}c@{\,~~~}c@{\,~~~}}
            \toprule
             \multicolumn{1}{c|}{\multirow{2}{*}{\textbf{Methods}}}& \multicolumn{5}{c|}{\textbf{MFNet Dataset}} & \multicolumn{5}{c|}{\textbf{FMB Dataset}} & \multicolumn{5}{c}{\textbf{LLVIP Dataset}} \\\cline{2-16}
              & MI & EN & CC & SSIM & $Q^{AB/F}$ &  MI & EN & CC & SSIM & $Q^{AB/F}$ &  MI & EN & CC & SSIM & $Q^{AB/F}$\\
            \midrule
            SDNet  & 1.325 & 5.827 & 0.592 & 0.861 & 0.456 & 2.285 & 6.617 & \textbf{0.580} & 0.926 & 0.540 & 1.535 & 6.965 &	0.692 &	0.831 &	0.527 \\
            U2Fusion  & 1.426 &	5.185 &	0.618 &	0.650 &	0.349 & 1.983 &	6.410 &	\underline{0.579} &	\underline{0.987} &	0.556 & 1.329 & 6.807 & 0.715 & 0.839 & 0.483 \\
            Tar-DAL  & 1.942 & 6.337 & 0.623 & 0.838 & 0.452 & 2.190 & 6.472 & 0.540 & 0.897 & 0.415 & 1.953 & 7.411 & 0.696 & 0.790 & 0.388 \\
            MetaFusion   & 1.214 & 6.049 & 0.592 & 0.672 & 0.401 & 1.617 & \underline{6.654} & 0.547 & 0.594 & 0.412 & 1.022 &	7.401 &	0.667 &	0.673 &	0.301 \\
            LRRNet   & 1.632 & 5.458 & 0.554 & 0.548 & 0.464 & 2.163 & 6.281 &	0.552 &	0.767 &	0.534 & 1.451 & 6.611 & 0.674 & 0.831 & 0.427 \\
            Diff-IF & 2.432 & 6.323 & 0.611 & 0.891 & \textbf{0.688} & \underline{2.745} & 6.623 & 0.507 & 0.963 & 0.639 & 2.185 & 7.455 &	0.702 &	0.904 &	0.598 \\
            EMMA   & \underline{2.540} & 6.353 & 0.617 & 0.908 & 0.601 & 2.725 & 6.520 & 0.527 & 0.914 & 0.630 & 2.137 & 7.441 &	\underline{0.716} &	\textbf{0.934} & 0.603 \\
            \rowcolor{blue!10}
            CDDFuse  & 2.172 & 6.309 & 0.610 & \textbf{0.973} & 0.626 & 2.710 & 6.651 & 0.557 & \textbf{0.988} & \textbf{0.657} & \underline{2.305} & 7.495 & 0.711 & \underline{0.927} & \underline{0.618} \\
            Text-IF   & 2.346 &	\underline{6.381} &	\underline{0.614} &	0.941 &	\underline{0.683} & 2.645 &	6.503 &	0.528 &	0.932 &	\underline{0.651} & 1.905 & \underline{7.536} & 0.704 & 0.910 & \textbf{0.651} \\
            \rowcolor{pink!40}
            \textbf{LUT-Fuse (ours)}   & \textbf{2.560} & \textbf{6.394} & \textbf{0.619} & \underline{0.966} & 0.628 & \textbf{2.999} & \textbf{6.662} & 0.507 & 0.906 &	0.613 & \textbf{2.446} &	\textbf{7.545} & \textbf{0.719} & 0.892 & {0.597} \\
            \midrule
            LUT-Fuse (MM-Net) & 2.764 & 6.425 & 0.615 & 0.971 & 0.706 & 3.014 & 6.671 & 0.588 & 0.927 & 0.643 & 2.502 & 7.581 & 0.705 & 0.946 & 0.746\\
            \bottomrule
        \end{tabular}
    }\label{quan_fusion}
\end{table*}

\noindent
\textbf{Datasets.} To validate the effectiveness of our proposed LUT-Fuse, we conducted comprehensive evaluations on publicly available infrared and visible image fusion datasets, including MFNet~\cite{ha2017mfnet}, FMB~\cite{liu2023multi}, and LLVIP~\cite{jia2021llvip}. The MFNet, FMB, and LLVIP datasets feature resolutions of 640$\times$480, 800$\times$600, and 1280$\times$1024, respectively. For our experiments, we utilized a total of 784 image pairs for training, while employing 150 images from MFNet, 100 from FMB, and 100 from LLVIP for testing purposes.

\noindent
\textbf{Metric.} We employ the metrics including the mutual information (MI)~\cite{qu2002information}, information entropy (EN)~\cite{roberts2008assessment}, correlation coefficient (CC), structural similarity index measure (SSIM)~\cite{wang2004image}, and quality of gradient-based fusion ($Q^{AB/F}$)~\cite{ma2019infrared}. Higher values of MI, EN, CC, SSIM, and $Q^{AB/F}$ indicate higher quality of the fusion image.

\noindent
\textbf{SOTA Competitors.} We compare LUT-Fuse with several state-of-the-art methods on multiple datasets. The methods for comparison include SDNet~\cite{zhang2021sdnet}, U2Fusion~\cite{xu2020u2fusion}, Tar-DAL~\cite{liu2022target}, MetaFusion~\cite{zhao2023metafusion}, LRRNet~\cite{li2023lrrnet}, Diff-IF\cite{yi2024diff}, EMMA~\cite{zhao2024equivariant}, CDDFuse~\cite{zhao2023cddfuse}, and Text-IF~\cite{yi2024text}.

\subsection{Qualitative Experiments} 
The qualitative results on multiple datasets are reported in Fig.~\ref{qual_fusion}. SDNet, U2Fusion, LRRNet poorly preserving thermal information and visible textures, could not obtain effective scene representation. Tar-DAL excels in infrared target object but fails in texture preservation, notably losing wheel details in the fourth row. CDDFuse, EMMA, and Text-IF achieve relatively good fusion results in most scenes, they fall behind LUT-Fuse in preserving human thermal radiation, as demonstrated in the first row. Overall, LUT-Fuse achieves remarkably competitive results in terms of highlighting thermal objects and clear visible texture while requiring only a fraction (around one-tenth) of the computational time compared to compared methods.

\subsection{Quantitative Experiments} 
The quantitative results on multiple datasets are reported in Tab.~\ref{quan_fusion}. Our proposed LUT-Fuse demonstrates comprehensively optimal fusion performance compared with state-of-the-art fusion methods in terms of MFNet, FMB, and LLVIP datasets. In terms of MI and EN, it outperforms the comparative methods. In terms of CC, SSIM, and $Q^{AB/F}$, it also gets competitive results. This demonstrates that LUT-Fuse maintains excellent quantitative performance while achieving exceptionally high computational speed.

\begin{figure}[!t]
\centering
\includegraphics[width=0.97\linewidth]{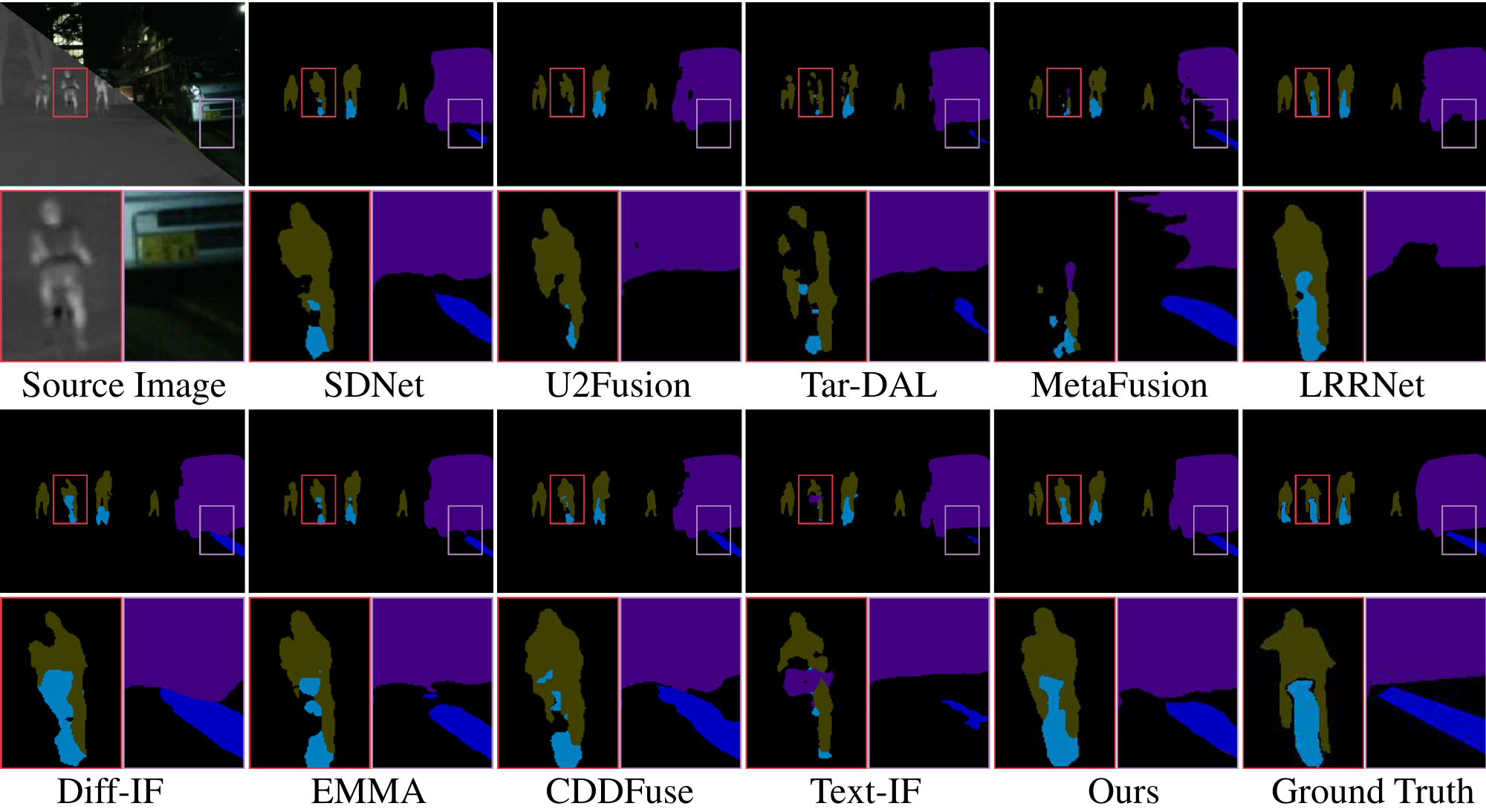}
\vspace{-0.1in}
\caption{Qualitative comparison in semantic segmentation task of our proposed LUT-Fuse with the state-of-the-art multi-modal image fusion methods on MFNet dataset.}
\label{seg_qual}
\end{figure}

\begin{figure}[!t]
\centering
\includegraphics[width=0.98\linewidth]{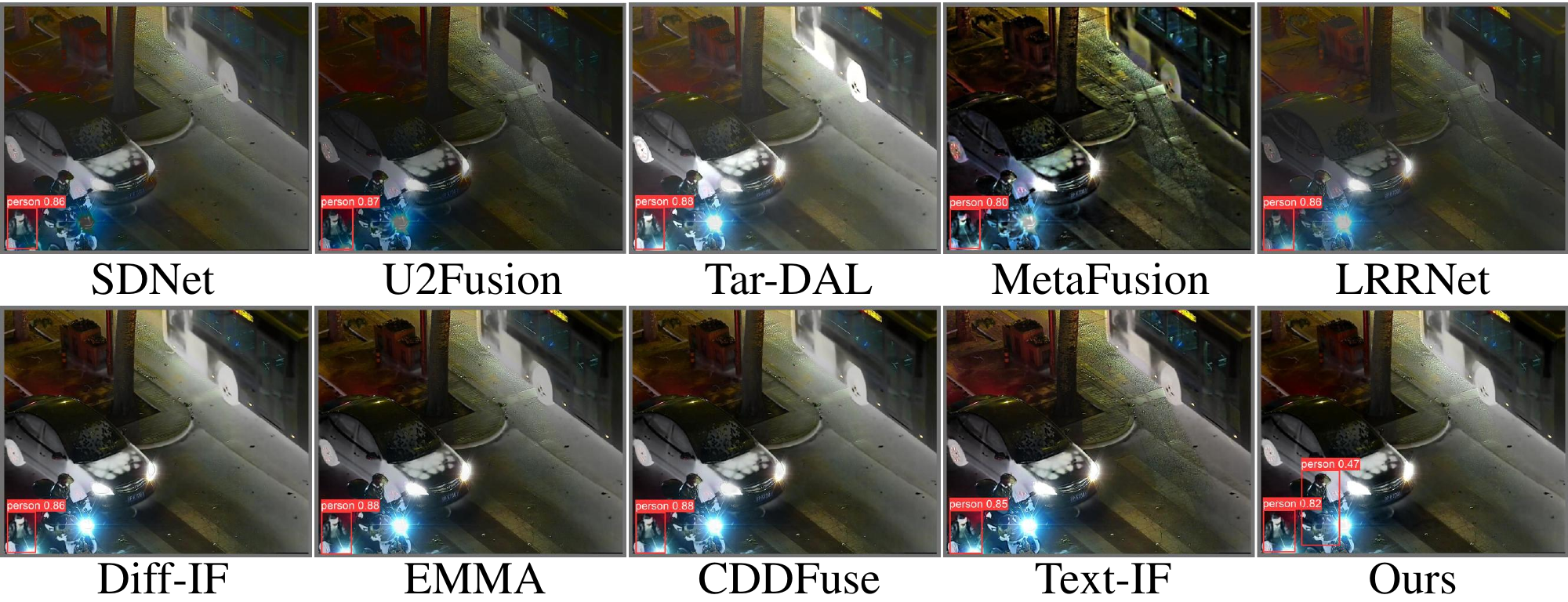}
\vspace{-0.1in}
\caption{Qualitative comparison in object detection task on LLVIP dataset. \textbf{Please zoom in for better viewing. }}
\label{det_qual}
\end{figure}

\begin{table*}[h]
    \centering
    \begin{minipage}{0.58\textwidth}
        \centering
        \caption{Quantitative comparison of semantic validation in MFNet dataset. (\textbf{Bold}: optimal performance, \underline{underline}: second-best performance)}
        \label{seg_quan}
    \vspace{-0.08in}
        \resizebox{1\textwidth}{!}{
        \begin{tabular}{c|*{9}c|c}
    \toprule
    \textbf{Methods} & Unlabel & Car & Person & Bike & Curve & Car Stop & Guard. & Cone & Bump & \textbf{mIoU} \\ \midrule
    SDNet & 98.05 & 86.99 & 72.83 & 62.22 & 43.45 & 18.53 & 3.95 & 50.37 & 55.94 & 54.70 \\
    U2Fusion & 97.94 & 85.81 & 71.28 & 62.62 & 37.16 & 31.56 & 7.05 & 44.10 & 53.84 & 54.59 \\
    Tar-DAL & 97.97 & 85.85 & 70.88 & 61.95 & 38.87 & 28.83 & 6.40 & 43.85 & 45.08 & 53.30 \\
    MetaFusion & 97.91 & 85.75 & 68.61 & 62.33 & 36.01 & 29.35 & 6.44 & 49.61 & 42.33 & 53.15 \\
    LRRNet & 98.02 & 86.43 & 71.78 & 62.92 & 39.20 & 30.39 & 8.56 & 44.63 & 49.48 & 54.60 \\
    Diff-IF & 97.94 & 84.63 & 71.55 & 61.90 & 41.74 & 18.83 & 4.81 & 50.07 & 47.42 & 53.21 \\
    EMMA & 98.04 & 87.30 & 72.28 & 62.37 & 44.35 & 30.40 & 6.91 & 44.44 & 45.88 & 54.66 \\
    CDDFuse & 98.05 & 86.33 & 72.32 & 61.34 & 41.94 & 21.37 & 3.20 & 49.15 & 48.08 & 53.53 \\ 
    \rowcolor{pink!40}
    Text-IF & 98.02 & 86.31 & 72.45 & 62.57 & 43.11 & 30.54 & 3.30 & 51.51 & 50.12 & \textbf{55.32} \\
    \rowcolor{blue!10}
    \textbf{LUT-Fuse (ours)} & 98.02 & 85.81 & 70.96 & 60.56 & 43.80 & 27.20 & 7.16 & 49.95 & 49.90 & \underline{54.82} \\
    \bottomrule
    \end{tabular}}
    \end{minipage}
    \hfill
    \begin{minipage}{0.35\textwidth}
        \centering
        \caption{Quantitative comparison of object detection in LLVIP dataset. }
        \label{det_quan}
    \vspace{-0.08in}
        \resizebox{1\textwidth}{!}{
        \begin{tabular}{c|ll|cc}
        \toprule
        \textbf{Methods} & Pre. & Rec. & mAP@0.5 & mAP@0.5:0.95 \\
        \midrule
        SDNet & 0.927 & \textbf{0.888} & 0.931 & 0.604 \\
        \rowcolor{blue!10}
        U2Fusion & \underline{0.947} & 0.874 & \textbf{0.943} & 0.609 \\
        Tar-DAL & 0.912 & 0.842 & 0.927 & 0.595 \\
        MetaFusion & 0.944 & 0.881 & 0.936 & 0.595  \\
        LRRNet & 0.930 & 0.873 & 0.937 & 0.605  \\
        Diff-IF & 0.936 & \underline{0.887} & 0.923 & \underline{0.611} \\
        EMMA & 0.938 & 0.851 & 0.932 & 0.605 \\
        CDDFuse & 0.933 & 0.867 & 0.927 & 0.599 \\
        Text-IF & 0.925 & 0.875 & 0.938 & 0.597 \\
        \rowcolor{pink!40}
        \textbf{LUT-Fuse (ours)} & \textbf{0.953} & 0.873 & \underline{0.941} & \textbf{0.614} \\
        \bottomrule
    \end{tabular}}
    \end{minipage}
\end{table*}

\subsection{Performance on High-Level Task}
To verify the performance in downstream high-level vision tasks, we conduct semantic segmentation and object detection experiments on MFNet and LLVIP, respectively.

\noindent
\textbf{Semantic Segmentation.} SegFormer~\cite{xie2021segformer} is adopted as the backbone in the semantic segmentation task. Qualitative and quantitative results are reported in Fig.~\ref{seg_qual} and Tab.~\ref{seg_quan}. In Fig.~\ref{seg_qual}, our method demonstrates optimal segmentation performance for both pedestrians and bicycles, while also showing highly competitive results in guardrail and car segmentation. In Tab.~\ref{seg_quan}, our method achieves the second-best performance, with only a marginal gap compared to Text-IF. This demonstrates that our approach maintains excellent semantic preservation capabilities while requiring minimal computational overhead.

\noindent
\textbf{Object Detection.} We employ YOLOv5\footnote{\url{https://github.com/ultralytics/yolov5}} as the object detection network and train it on the LLVIP dataset. Qualitative and quantitative experimental results are shown in Fig.~\ref{det_qual} and Tab.~\ref{det_quan}. In Fig.~\ref{det_qual}, particularly in scenarios where other methods suffer from missed detections, our method shows superior detection performance. In Tab.~\ref{det_quan}, LUT-Fuse also exhibits comprehensively optimal semantic detection performance, demonstrating its robustness representation.

\subsection{Ablation Experiments}
\begin{figure}[t]
\centering
\includegraphics[width=0.99\linewidth]{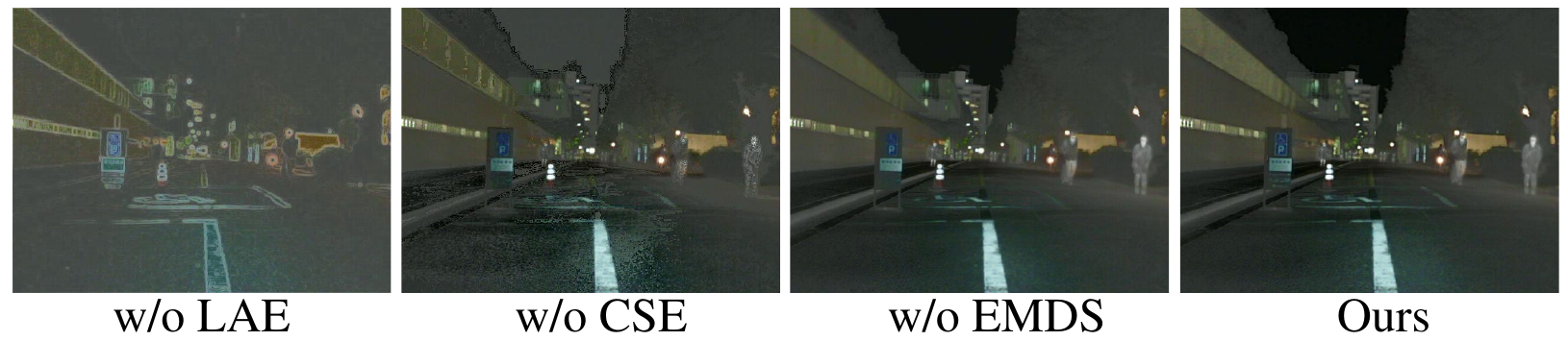}
\vspace{-0.1in}
\caption{Qualitative comparison of ablation experiment on MFNet dataset.}
\label{fig:ablation}
\end{figure}

To verify the effectiveness of the proposed module, we conduct ablation experiments on MFNet dataset. It includes the ablation of Low-order approximation encoding (LAE), high-level joint contextual scene encoding (CSE), and efficient MM-LUT distillation strategy (EMDS). Qualitative and quantitative results are presented in Fig.~\ref{fig:ablation} and Tab.~\ref{tab:ablation}.

In Fig.~\ref{fig:ablation}, it reveals that the removal of any individual module significantly degrades the fusion performance, demonstrating the essential contribution of each component to the overall system functionality. Also, our full model achieves the best quantitative fusion metrics, as clearly demonstrated in Tab.~\ref{tab:ablation}.

\begin{table}[t]
    \centering
    \setlength{\tabcolsep}{10pt}
    \renewcommand{\arraystretch}{1.08}
    \footnotesize
    \caption{Quantitative results of the ablation experiment. (\textbf{Bold} shows the optimal performance.)}
    \vspace{-0.08in}
    \resizebox{0.42\textwidth}{!}{
    \begin{tabular}{p{0.005\textwidth}p{0.005\textwidth}c|p{0.008\textwidth}p{0.008\textwidth}p{0.008\textwidth}p{0.008\textwidth}c}
        \toprule
        \textit{LAE} & \textit{CSE} & \textit{EMDS} & MI & EN & CC & SSIM & $Q^{AB/F}$\\
        \midrule
         & \checkmark  & \checkmark  & 1.812 & \textbf{6.574} & 0.491 & 0.533 & 0.407  \\
        \checkmark & & \checkmark & 1.747 & 6.445 & 0.508 & 0.553 & 0.404 \\
        \checkmark & \checkmark &  & 2.552 & 6.381 & 0.571 & 0.904 & 0.566 \\
        \checkmark & \checkmark & \checkmark & \textbf{2.560} & 6.394 & \textbf{0.619} & \textbf{0.966} & \textbf{0.628} \\
        \bottomrule
    \end{tabular}}
    \label{tab:ablation}
\end{table}

\subsection{Extended Experiments}

\begin{figure}[!t]
\centering
\includegraphics[width=0.99\linewidth]{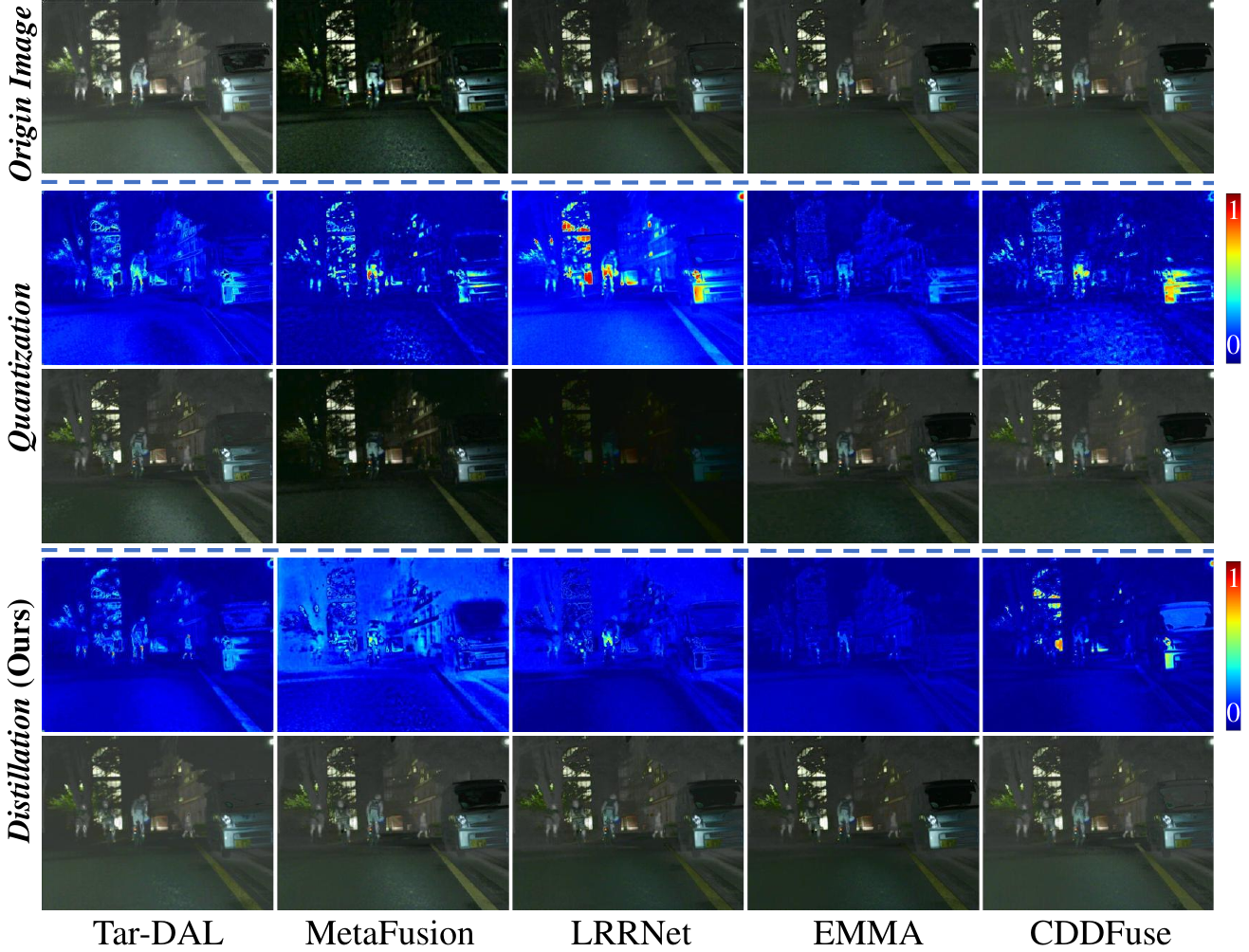}
\vspace{-0.1in}
\caption{Qualitative comparison of non-learnable quantization LUT and our proposed efficient learnable distillation MM-LUT.}
\label{fig:quan_dist_to_LUT}
\end{figure}

\begin{table}[!t]
    \centering
    \setlength{\tabcolsep}{8pt}
    \renewcommand{\arraystretch}{1.08}
    \footnotesize
    \centering
    \caption{Quantitative comparison of conventional non-learnable quantization (\textit{quanti.}) and learnable distillation (\textit{dist.}) to LUT in MFNet dataset. (\textbf{Bold}: shows the optimal performance.)}
    \label{tab:quan_dist_to_LUT}
    \vspace{-0.08in}
    \resizebox{0.48\textwidth}{!}{
    \begin{tabular}{c|c|ccccc}
        \toprule
        \textbf{Methods} & \textbf{Type} & MI & EN & CC & SSIM & $Q^{AB/F}$\\
        \midrule
        \multirow{2}{*}{\textbf{Tar-DAL}} & \textit{quanti.} & 1.236 & 4.995 & 0.408 & 0.476 & 0.306\\
         & \cellcolor{pink!40}\textit{dist.(Ours)} & \cellcolor{pink!40}\textbf{1.713} & \cellcolor{pink!40}\textbf{6.017} & \cellcolor{pink!40}\textbf{0.585} & \cellcolor{pink!40}\textbf{0.745} & \cellcolor{pink!40}\textbf{0.363} \\
        \midrule
        \multirow{2}{*}{\textbf{MetaFusion}} & \textit{quanti.} & 1.127 & 5.355 & 0.468 & 0.559 & 0.354\\
        & \cellcolor{pink!40}\textit{dist.(Ours)} & \cellcolor{pink!40}\textbf{1.152} & \cellcolor{pink!40}\textbf{5.817} & \cellcolor{pink!40}\textbf{0.553} & \cellcolor{pink!40}\textbf{0.636} & \cellcolor{pink!40}\textbf{0.375} \\
        \midrule
        \multirow{2}{*}{\textbf{LRRNet}} & \textit{quanti.} & 1.174 & 4.698 & 0.405 & 0.298 & 0.265\\
        & \cellcolor{pink!40}\textit{dist.(Ours)} & \cellcolor{pink!40}\textbf{1.542} & \cellcolor{pink!40}\textbf{5.301} & \cellcolor{pink!40}\textbf{0.547} & \cellcolor{pink!40}\textbf{0.473} & \cellcolor{pink!40}\textbf{0.401} \\
        \midrule
        \multirow{2}{*}{\textbf{Diff-IF}} & 
        \textit{quanti.} & 2.001 & 5.251 & 0.568 & 0.513 & 0.202 \\
        & \cellcolor{pink!40}\textit{dist.(Ours)} & \cellcolor{pink!40}\textbf{2.251} & \cellcolor{pink!40}\textbf{6.313} & \cellcolor{pink!40}\textbf{0.602} & \cellcolor{pink!40}\textbf{0.817} & \cellcolor{pink!40}\textbf{0.625} \\
        \midrule
        \multirow{2}{*}{\textbf{EMMA}} & 
        \textit{quanti.} & 1.903 & 6.219 & 0.518 & 0.725 & 0.513 \\
        & \cellcolor{pink!40}\textit{dist.(Ours)} & \cellcolor{pink!40}\textbf{2.374} & \cellcolor{pink!40}\textbf{6.275} & \cellcolor{pink!40}\textbf{0.593} & \cellcolor{pink!40}\textbf{0.815} & \cellcolor{pink!40}\textbf{0.552} \\
        \midrule
        \multirow{2}{*}{\textbf{CDDFuse}} & 
        \textit{quanti.} & 1.733 & 5.125 & 0.513 & 0.553 & 0.478 \\
        & \cellcolor{pink!40}\textit{dist.(Ours)} & \cellcolor{pink!40}\textbf{2.087} & \cellcolor{pink!40}\textbf{6.128} & \cellcolor{pink!40}\textbf{0.595} & \cellcolor{pink!40}\textbf{0.857} & \cellcolor{pink!40}\textbf{0.568} \\
        \bottomrule
    \end{tabular}
    }
    \label{tab:lut_kd}
\end{table}

\textbf{Quantization vs. Distillation MM-LUT.} 
Our proposed LUT-accelerated MMIF strategy demonstrates broad applicability across various fusion backbones. Although non-learnable quantization LUT methods, currently the mainstream approach~\cite{jiang2023meflut}, can do this to some extent by simulating input data and directly storing the network model output in LUTs. However, this strategy inevitably leads to reduced LUT precision and poor generalization capabilities. In contrast, our proposed learnable LUT framework leverages efficient LUT distillation to directly optimize the look-up table parameters, achieving superior performance. We conduct a series of experiments in SOTA methods. Both qualitative and quantitative results are presented in Fig.~\ref{fig:quan_dist_to_LUT} and Tab.~\ref{tab:quan_dist_to_LUT}.

As indicated in residual maps of Fig.~\ref{fig:quan_dist_to_LUT}, our proposed distillation MM-LUT presents lower error compared to quantization solutions. In Tab.~\ref{tab:quan_dist_to_LUT}, our method achieves significant metric improvements across almost all experimental evaluations, demonstrating its better performance.

\begin{table}[t]
    \centering
    \setlength{\tabcolsep}{8pt}
    \renewcommand{\arraystretch}{1.2}
    \centering
    \caption{Running time of SOTA multi-modal image fusion methods and LUT-Fuse on \textbf{NVIDIA GeForce RTX 4060 Ti}. (\textcolor{green}{\ding{51}}: yes, \textcolor{orange}{$\sim$}: partially supported, \textcolor{red}{\ding{55}}: no)}
    \label{PC_plat}
    \vspace{-0.08in}
    \resizebox{0.492\textwidth}{!}{
    \begin{tabular}{c|>{\centering\arraybackslash}p{2.5cm}@{\,~~~}|>{\centering\arraybackslash}p{2.5cm}@{\,~~~}|>{\centering\arraybackslash}p{2.5cm}@{\,~~~}|c}
        \toprule
        \multirow{2}{*}{\textbf{Methods}} & \textbf{MFNet} & \textbf{FMB} & \textbf{LLVIP} & \multirow{2}{*}{Real Time}\\
        \cline{2-4}
        & Time/ms & Time/ms & Time/ms \\
        \midrule
        SDNet & 35.41 $\pm$ 8.23 & 40.14 $\pm$ 7.78 & 72.04 $\pm$ 7.99 & \textcolor{orange}{$\sim$}\\
        U2Fusion  & 64.24 $\pm$ 1.21 & 98.50 $\pm$ 0.51 & 268.67 $\pm$ 2.85 & \textcolor{red}{\ding{55}}\\
        Tar-DAL & 21.83 $\pm$ 0.34 & 36.08 $\pm$ 0.36 & 102.93 $\pm$ 1.07 & \textcolor{orange}{$\sim$}\\
        MetaFusion & 96.51 $\pm$ 1.20 & 97.43 $\pm$ 1.80 & 98.19 $\pm$ 2.34 & \textcolor{red}{\ding{55}}\\
        LRRNet & 314.85 $\pm$ 3.79 & 505.37 $\pm$ 4.28 & 1357.12 $\pm$ 5.50 & \textcolor{red}{\ding{55}}\\
        Diff-IF & 1723.05 $\pm$ 16.86 & 2818.82 $\pm$ 24.52 & 8800.50 $\pm$ 51.09 & \textcolor{red}{\ding{55}} \\
        EMMA & 126.18 $\pm$ 8.91 & 163.58 $\pm$ 6.54 & 493.61 $\pm$ 24.09 & \textcolor{red}{\ding{55}} \\
        CDDFuse & 641.13 $\pm$ 5.62 & 1045.06 $\pm$ 10.47 & 2791.07 $\pm$ 27.47 & \textcolor{red}{\ding{55}}\\
        Text-IF & 522.38 $\pm$ 2.55 & 826.96 $\pm$ 10.28 & 2491.64 $\pm$ 36.10 &  \textcolor{red}{\ding{55}} \\
        \textbf{Ours} & 4.70 $\pm$ 0.70 & 8.20 $\pm$ 1.80 & 23.20 $\pm$ 0.90 & \textcolor{green}{\ding{51}} \\
        \bottomrule
    \end{tabular}
    }
\end{table}

\subsection{Running Time \& Deployment Experiments} 
In practical applications, real-time performance is crucial for algorithm usability. The primary advantage of LUT-Fuse lies in its exceptional computational speed combined with competitive fusion quality. We validate this from two perspectives: PC and mobile/edge device platforms.

\noindent
\textbf{PC Platform.} In the NVIDIA GeForce RTX 4060 Ti platform, as shown in Tab.~\ref{PC_plat}, even methods specifically designed for real-time fusion can achieve only quasi-real-time performance in limited scenarios. This indicates that most existing methods struggle to meet real-time requirements even when deployed on high-performance computing platforms like GeForce RTX 4060 Ti with a power consumption of 165W, highlighting significant limitations in their practical applicability to real-world scenarios. Our proposed LUT-Fuse stands as the only method achieving real-time, and even super-real-time performance across all datasets.

\noindent
\textbf{Mobile Device Platform.} In the NVIDIA Jetson Orin NX platform, as reported in Tab.~\ref{edge_platform}, all comparative methods, including existing approaches specifically designed for real-time fusion, fail to achieve real-time performance on mobile processing devices. By contrast, our LUT-Fuse maintains real-time performance, which is the only one that can achieve this, demonstrating its adaptability to various practical applications. Compared with the current state-of-the-art lightweight methods (such as Tar-DAL in 720P), LUT-Fuse typically requires only about one-tenth of their computational time. Therefore, the ability to maintain real-time processing speeds on edge and mobile devices stands as a particularly distinctive feature of LUT-Fuse.

\begin{table}[!t]
    \centering
    \setlength{\tabcolsep}{8pt}
    \renewcommand{\arraystretch}{1.2}
    \centering
    \caption{Running time of SOTA multi-modal image fusion methods and LUT-Fuse on \textbf{NVIDIA Jetson Orin NX}. (\textcolor{green}{\ding{51}}: yes, \textcolor{red}{\ding{55}}: no)}
    \label{edge_platform}
    \vspace{-0.08in}
    \resizebox{0.492\textwidth}{!}{
    \begin{tabular}{c|c>{\centering\arraybackslash}p{1.5cm}@{\,~~~}|c>{\centering\arraybackslash}p{1.5cm}@{\,~~~}}
        \toprule
        \multirow{2}{*}{\textbf{Methods}} & \multicolumn{2}{c|}{480P (640 $\times$ 480)} & \multicolumn{2}{c}{720P (1280 $\times$ 720)} \\
        \cline{2-3} \cline{4-5}
        & Time/ms & Real Time & Time/ms & Real Time \\
        \midrule
        SDNet & 126.50 $\pm$ 2.86 & \textcolor{red}{\ding{55}} & 376.66 $\pm$ 4.00 & \textcolor{red}{\ding{55}}\\
        U2Fusion  & 249.23 $\pm$ 3.27 & \textcolor{red}{\ding{55}} & 778.42 $\pm$ 45.33 & \textcolor{red}{\ding{55}}\\
        Tar-DAL & 114.21 $\pm$ 3.15 & \textcolor{red}{\ding{55}} & 389.34 $\pm$ 40.08 & \textcolor{red}{\ding{55}}\\
        MetaFusion & 484.41 $\pm$ 4.32 & \textcolor{red}{\ding{55}} & 1542.38 $\pm$ 27.04 & \textcolor{red}{\ding{55}}\\
        LRRNet & 1368.37 $\pm$ 9.59 & \textcolor{red}{\ding{55}} & 3988.30 $\pm$ 99.85 & \textcolor{red}{\ding{55}}\\
        Diff-IF & 8200.11 $\pm$ 398.25 & \textcolor{red}{\ding{55}} & 29447.30 $\pm$ 2372.92 & \textcolor{red}{\ding{55}}\\
        EMMA & 459.61 $\pm$ 17.14 & \textcolor{red}{\ding{55}} & 2127.95 $\pm$ 43.99 & \textcolor{red}{\ding{55}}\\
        CDDFuse & 2630.50 $\pm$ 48.64 & \textcolor{red}{\ding{55}} & 8209.51 $\pm$ 170.11 & \textcolor{red}{\ding{55}}\\
        Text-IF & 2894.27 $\pm$ 239.80 & \textcolor{red}{\ding{55}} & 7750.53 $\pm$ 24.56 & \textcolor{red}{\ding{55}}\\
        \textbf{Ours}  & 18.23 $\pm$ 1.62 &  \textcolor{green}{\ding{51}}  & 30.54 $\pm$ 2.22 & \textcolor{green}{\ding{51}} \\
        \bottomrule
    \end{tabular}
    }
\end{table}

\section{Conclusion}
In this paper, we proposed a novel LUT framework towards extremely fast infrared and visible image fusion, termed as LUT-Fuse, which is the first application of LUTs in multi-modal image fusion to the best of our knowledge. It consists of a learnable LUT that is equipped with low-order approximation encoding and high-level joint contextual scene encoding, which is well-suited for multi-modal fusion. Given the lack of ground truth in MMIF, we naturally propose the efficient MM-LUT distillation strategy instead of traditional quantization LUT methods. It requires only around typically one-tenth of the time compared to current SOTA fusion algorithms, ensuring real-time performance even in low-power mobile devices. Extensive experiments validate the superiority, reliability, and stability of our proposed approach. In future research, our method shows strong potential for various fusion tasks, paving the way for advances in real-time image fusion.

\section*{Acknowledgments}
This work was supported by NSFC (62276192).

{
    \small
    \bibliographystyle{ieeenat_fullname}
    \bibliography{main}
}

\end{document}